\documentclass[letterpaper]{article}

\usepackage{natbib,alifeconf}  
\usepackage{amsfonts}
\usepackage{amsmath}
\usepackage{amssymb}
\usepackage{hyperref} 
\usepackage{url}
\DeclareMathOperator*{\argmax}{arg\,max}

\title{Active Dynamical Prospection: Modeling Mental Simulation as Particle Filtering for Sensorimotor Control during Pathfinding}

\author{Jeremy Gordon \and John Chuang \\
\mbox{}\\
University of California, Berkeley, Berkeley, CA 94704 \\
jrgordon@berkeley.edu} 

\begin{document}
\maketitle

\begin{abstract}

What do humans do when confronted with a common challenge: we know where we want to go but we are not yet sure the best way to get there, or even if we can. This is the problem posed to agents during spatial navigation and pathfinding, and its solution may give us clues about the more abstract domain of planning in general. In this work, we model pathfinding behavior in a continuous, explicitly exploratory paradigm. In our task, participants (and agents) must coordinate both visual exploration and navigation within a partially observable environment. Our contribution has three primary components: 1) an analysis of behavioral data from 81 human participants in a novel pathfinding paradigm conducted as an online experiment, 2) a proposal to model prospective mental simulation during navigation as particle filtering, and 3) an instantiation of this proposal in a computational agent. We show that our model, Active Dynamical Prospection, demonstrates similar patterns of map solution rate, path selection, and trial duration, as well as attentional behavior (at both aggregate and individual levels) when compared with data from human participants. We also find that both distal attention and delay prior to first move (both potential correlates of prospective simulation) are predictive of task performance.
\end{abstract}


\section{Introduction}

What do humans do when confronted with a common challenge: we know where we want to go (perhaps we can even see our destination already), but we are not yet sure the best way to get there, or even if we can. This is the problem posed to agents during spatial navigation and pathfinding, and its solution may give us clues that extend into the more abstract domain of planning in general.

In this work, we aim to analyze and model pathfinding behavior in a task paradigm that is more continuous and dynamic than those historically chosen by the planning literature. In our task, participants (and agents) must coordinate both visual exploration and navigation within a partially observable environment in which the dynamics of movement result in ongoing uncertainty about the true passability of potential paths.

This contribution has three primary components: 1) an analysis of behavioral data from a novel pathfinding task conducted as an online experiment, 2) a proposal to model mental simulation during navigation as particle filtering, and 3) an instantiation of this proposal in an agent capable of solving the task in ways that share attributes with human performance.

By developing a computational model of active perception, simulation and movement during our pathfinding task, and comparing results with human behavioral data, we hope to shed light on the following questions:

\begin{itemize}
    \itemsep -0.3em
    \item How are simulations of potential future actions coordinated during pathfinding and navigation?
    \item Which path characteristics attract attention and forward simulation?
    \item What are the distributional and temporal dynamics of attention, and how do they relate with pathfinding performance?
    \item Can a common computational mechanism successfully drive the coordination of both visual attention and navigation?
\end{itemize}

\subsection{Background \& Related Models}

\subsubsection{Situated Planning}

An extensive literature exists around planning across cognitive science, psychology, neuroscience, and artificial intelligence. Often, planning problems are posed in line with classical problem solving, in which the environment is fully observed with known dynamics (e.g. formalized as a Markov Decision Process), and solution entails identifying a sequence of actions resulting in a goal condition \citep{newell1972human}. In reinforcement learning, Monte Carlo methods are frequently used to sample trajectories during value estimation, and therefore to support the planning of future actions. \cite{silver2010monte} proposed Partially Observable Monte Carlo Planning (POMCP) to make value estimation tractable in high dimensional state spaces. In this work, particle filtering is used to efficiently approximate belief state updates when access to the true generative process is not available. 

In embodied planning, agents are situated within complex, noisy, and uncertain environments in which, importantly, they must control both sensors and other motor outputs while simultaneously planning future actions in an online fashion. Though common for some time in robotics, efforts to develop theories of realistic embodied planning have recently gained momentum, propelled by multidisciplinary contributions from dynamical systems, ecological psychology, and reinforcement learning.

To select just a few examples, \cite{cos2021changes} demonstrated that perturbations to the arm during a reaching task can prompt changes of mind, indicating that deliberation continues dynamically during action execution. \cite{pezzulo2019planning} proposed a connection between specific neural dynamics (sharp-wave ripples and theta sequences) as mechanisms to support planning in two regimes: at decision time, and in the background to optimize a behavioral controller. In a foraging paradigm, \cite{yoon2018control} developed a model of normative utility based on the marginal value theorem, and applied it to a visual information harvesting experiment in which fixation duration (time spent at a patch) and saccade speed (movement vigor between patches) were measured. Their findings suggest a shared principle of control may underlie both aspects of foraging behavior.

\subsubsection{Navigation, Simulation \& Prospection}

According to \cite{montello2005navigation}, navigation can be decomposed into two components: 1) locomotion, in which the body is coordinated to its local surrounds, and 2) wayfinding, in which a goal-directed agent plans actions aided by memory of both the local and distal environment. Though a range of neuroscientific mechanisms have been proposed to support both components (e.g. cells in the hippocampal formation encoding position, orientation, and head direction, among others), the dynamics by which internal models of the environment are queried offline (via simulation) and integrated with present sensory information (e.g. the observation of landmarks), is not well understood.

Mental simulation, often also referred to as replay or preplay, is the generation of internal sequences reflecting previous or possible engagements with the world. In a psychophysics experiment, \cite{arnold2016mental} showed that humans adaptively compress simulations of potential routes during prospective route planning. \cite{chersi2013mental} developed a computational model of simulated and overt action during maze navigation, involving the hippocampus and striatum to support recall and cache action values respectively.

Especially important to the present work, is active inference, which has been proposed as a suitable formal theory to support the project of embodied perception, action, and planning \citep{friston2017active}. In active inference, agents act to reduce prediction error (free energy) produced from inconsistencies between an internal generative model and sensory observations. Recent work has applied the theory to planning and navigation, running simulations of navigation in a maze environment \citep{kaplan2018planning}. 

Finlly, the literature on active navigation has investigated the relationship between sensory exploration and pathfinding. In a recent study, \cite{lakshminarasimhan2020tracking} showed that eye movements could be used to infer latent beliefs such as the location of a hidden goal during virtual spatial navigation, and that controlling fixations had detrimental effects on navigation performance.




\subsubsection{Swarm Intelligence}

Our model also draws inspiration from \cite{trianni2009swarm}, who argued that the integration of artificial life and cognitive science via ``swarm cognition'' could offer fruitful progress in understanding and modeling cognitive mechanisms. Swarm intelligence has demonstrated that simple unit-level behaviors (such as individual particle dynamics), when operating as a collective system, can produce complex emergent properties. In some cases, such as that of ant colony optimization, these system-level properties offer strategies effective at even NP-hard problems, such as the traveling salesman \citep{colorni1991distributed}.  


\section{Online Pathfinding Experiment}

\begin{figure}
\begin{center}
\includegraphics[width=\linewidth]{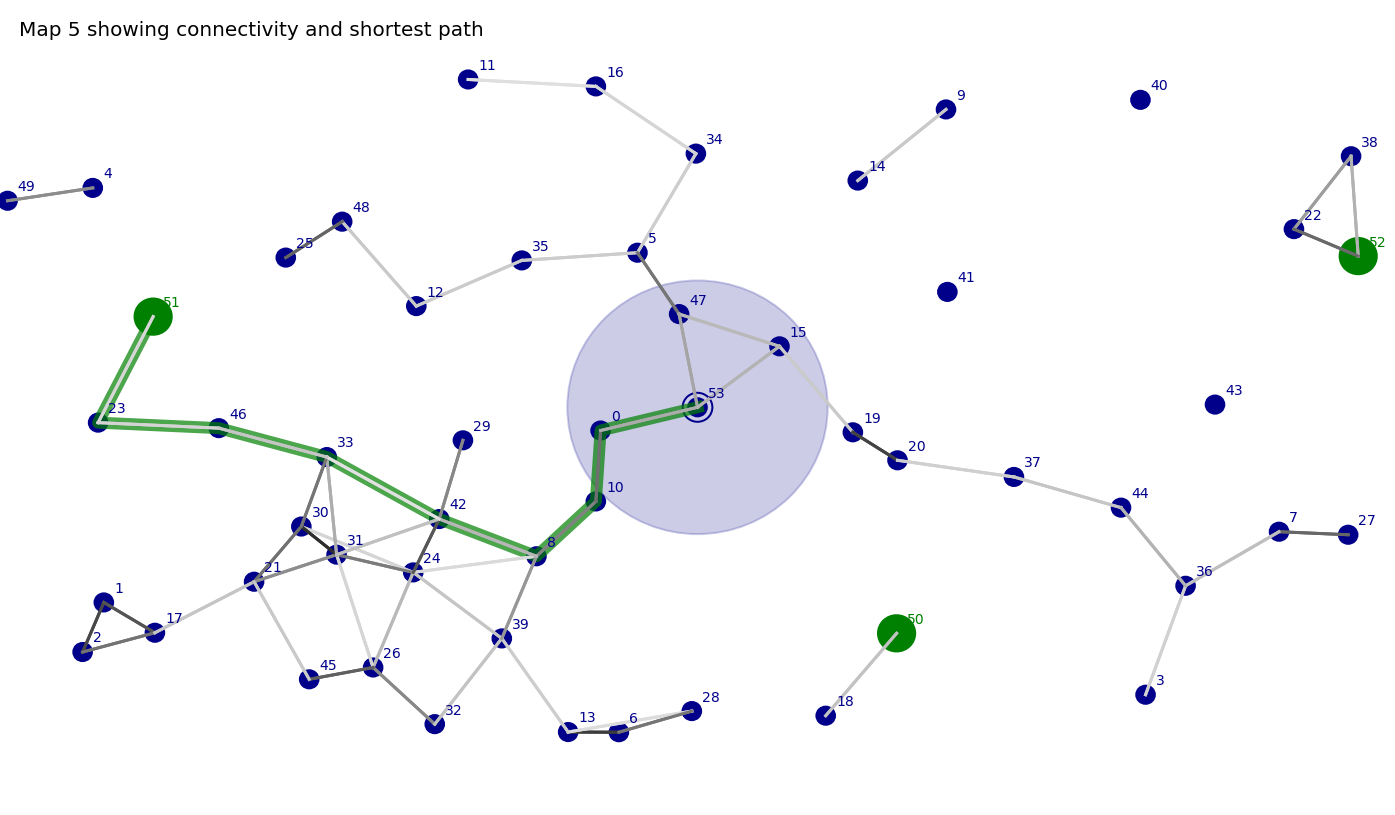}
\caption{Sample map (map 5) with hold connectivity plotted as edges (lighter edges indicate gaps closer to reach limit). Optimal path to goal is plotted in green. The transparent blue circle indicates the reach radius, at the center of which, the small blue ring indicates the reach target at the agent's present location.} \label{fig:map_example}
\end{center}
\end{figure}

Our task was designed to require participants to explicitly coordinate visual attention and navigation to a goal. On each trial, participants saw their present location (at the center of the screen), as well as the locations of one or more goals. A landscape of 50 `holds' was initially hidden, and exposed when the particpant moved their cursor across the landscape, exploring the map in a spotlight-like manner. Holds were reachable only when within a fixed radius of the participant's present position (indicated by a blue circle as shown in Figure \ref{fig:map_example}). To navigate to a reachable hold, the participant dragged it toward the small central target indicating their present location. During a successful drag, the full landscape shifted such that the chosen hold became centered within the egocentric space. In this way the participant was able to navigate towards and eventually reach a chosen goal.

By designing the task in this partially observable, egocentric manner, we were able to capture both movement and attention independently, and ensure that computational models of behavior in this paradigm contend with the richness of online sensorimotor exploration. 



\subsection{Participants}

Study participants were recruited through an on-campus experimental lab at a public university in the United States.  

81 participants completed the online study. Participants had a mean age of 22 $\pm$ 2.1. 61 identified themselves as women, 18 as men, 1 as non-binary or non-conforming, and 1 declined to answer. 56 reported their race as Asian, 13 white, 1 Black or African American, 1 American Indian or Alaska Native, and 7 Other, including White and Asian (2) and Middle Eastern (2). 9 participants identified as Spanish, Hispanic, or Latino. All participants were undergraduate students, graduate students or staff at the University of California, Berkeley.

\subsection{Procedure}


The experiment began with a series of instructions about the task. Participants completed a practice trial where they were guided through a trivial landscape to a nearby goal location to ensure they understood the mechanics of navigation and the trial objective. The full experiment entailed completing each of 11 predefined maps in randomized order. Each trial ended when a goal was reached, or when a 60-second trial timer expired. Pariticipants received a base incentive of \$6, and a performance bonus of \$0, \$2, or \$4 depending on final score as a percent of maximum (less than 60\%, 60-80\%, or more than 80\%). The recruitment process and study protocol were approved by the local ethics review board.

\subsection{Data Structure}
\label{sec:data_structure}

Two types of data were captured during each trial: navigation data, and `attention' data. Navigation data included each attempt to navigate to a hold in the landscape, whether successful or unsuccessful, resulting in a final path through the landscape represented as a list of holds and timestamps. Attention data was recorded as a stream of 2D cursor coordinates $(x, y)$ captured at 30 Hz. 



\begin{figure}
\begin{center}
\includegraphics[width=\linewidth]{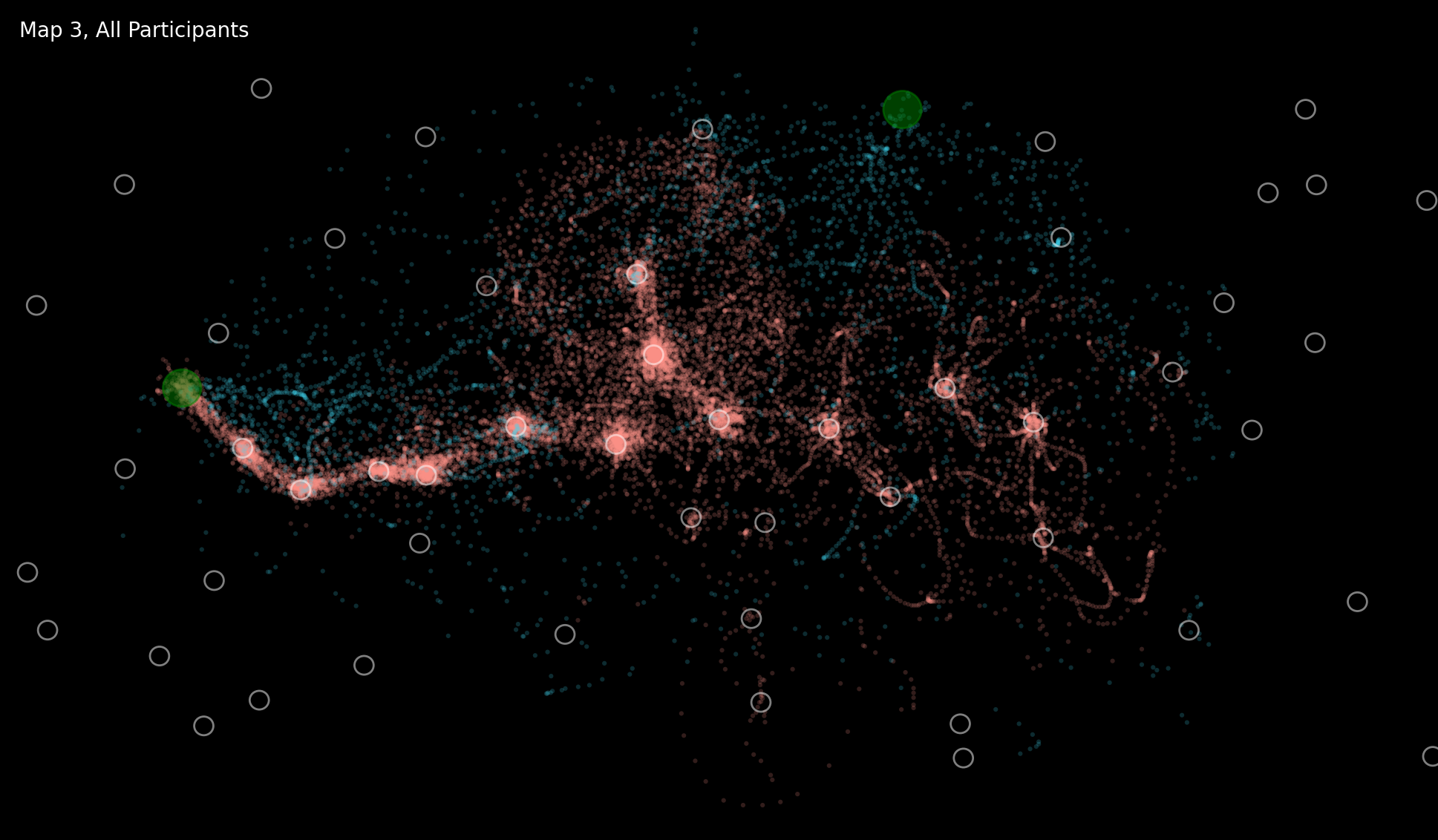}
\caption{Plot of attention across all participants (map 3). Red points indicate attention within reach zone. Blue points indicate attention beyond reach zone, a proxy for exploration.} \label{fig:sample_map_attention}
\end{center}
\end{figure}


Videos rendering all participants' navigation and attention data for all maps are available on the first author's website~\footnote{All media is available at: \url{https://jgordon.io/project/adp}}.

\subsection{Behavioral Data Analysis}
\label{sec:behavioral_data_analysis}

To compare performance metrics with map difficulty, and given the small number of maps in our dataset, we elected to define three difficulty categories (low, medium, and high-difficulty maps) based on the sample-wide success rate across all participants. In addition, we extracted a number of behavioral metrics from the raw navigation and attention data.

\subsubsection{Prospection and Spatial Exploration}

By exploiting one feature of the task design---that cursor coordinates farther from the origin indicate visual exploration of more distant, not presently reachable, holds---we defined a key metric, attention distance, as the Euclidean distance of the cursor (fovea) position from the agent (at screen center). Furthermore, by segmenting attentional coordinates as shown in Figure \ref{fig:sample_map_attention}~into \emph{reachable} (red) and \emph{unreachable} (blue) groups, spatial patterns relating to map exploration could be directly visualized.



\subsubsection{Trial Score}

In order to compare the computed behavioral metrics to an indication of both trial-level and participant-level performance, we define the score $\sigma_{ij}$ for participant $i$ and map $j$ via the function: 

\begin{equation*}
\sigma_{ij} = f(path_{ij}, map_j) = \begin{cases}
          0 \quad &\text{if} \, success = 0 \\
          \frac{\lambda_{min}}{\lambda_{pp}} \quad &\text{if} \, success = 1 \\
     \end{cases}
\end{equation*}

Here, $\lambda_{pp}$ was computed as the number of successful moves completed by the participant, and $\lambda_{min}$ was a map-specific property defined as the minimum-length path to (any) goal. As such, $\sigma_{ij} \in [0, 1]$.

\section{Computational Model}

\subsection{Model Rationale}


We propose Active Dynamical Prospection (ADP), a model of planning related to active inference and augmented by ideas from swarm intelligence and dynamical systems. Following active inference, we assume that mental simulation may be leveraged to simultaneously learn, and plan within, a generative model of the agent's environment. 

Our computational model is guided by the following central hypothesis: that covert mental simulations supporting this task may be fruitfully modeled as Monte Carlo particle filtering across a learned energy landscape, and subject to a set of precise physical dynamics aligned with the interaction capabilities of the agent within its environment. While \cite{tschantz2020scaling} discusses the use of trajectory sampling to learn the generative density in active inference, what we propose is a stronger commitment to particle filtering as a descriptive model of simulation with possible links to covert attention.

\begin{figure}[t]
\begin{center}
\includegraphics[width=\linewidth]{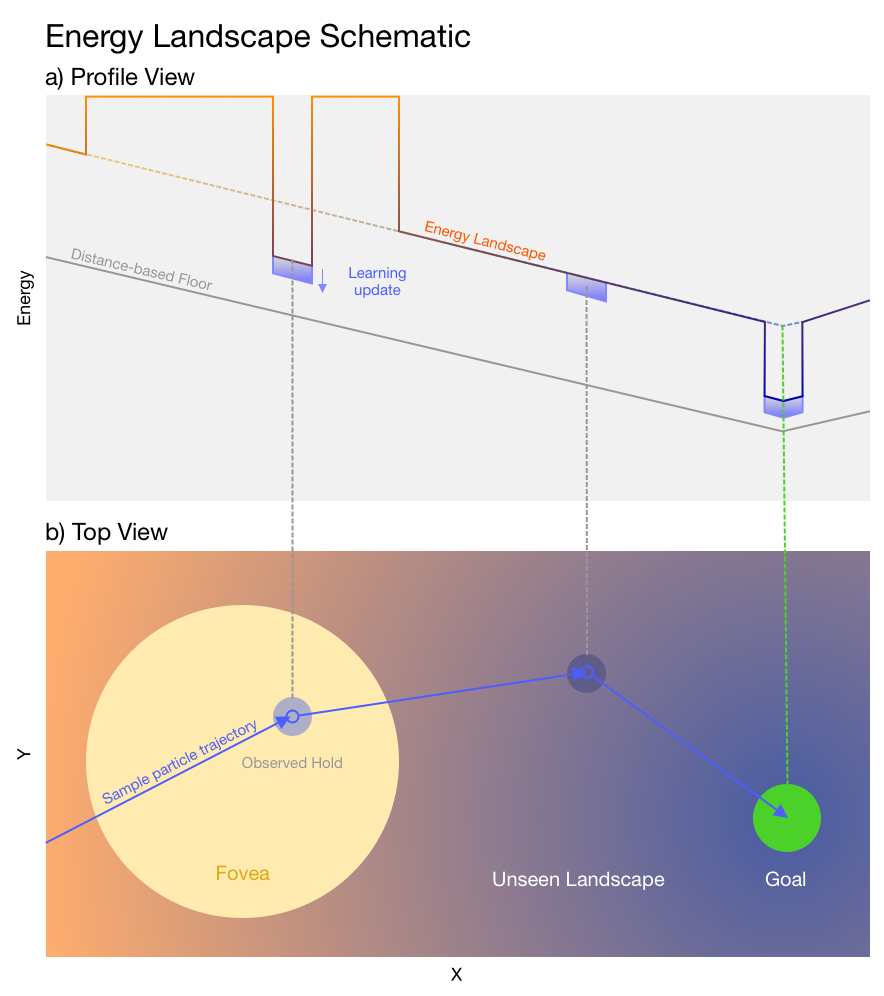}
\caption{Energy landscape schematic (profile and top-down view of sample map region). Particle trajectories are influenced by an energy well (darker blue) in the landscape around the goal location. The rollout shown passes through the high energy location under the fovea (yellow) only because of the hold observed there.} \label{fig:1d_energy}
\end{center}
\end{figure}

We view pathfinding as prospective inference, or the act of reducing uncertainty over the ultimate trajectory an agent will take through its environment. In our model, the agent learns a representation of the movement affordances in its environment, which can be thought of as a 2-dimensional energy surface. Given initial visual access only to its own location and that of the goal(s), agents begin with a sensible, but na\"ive, prior form for this surface, which we model as distance to nearest goal (see the gradient surrounding the goal in Figure \ref{fig:1d_energy}b).

Agents leverage a set of three tools to uncover the true topology of their environment: 1) overt visual search by moving the fovea to expose hold locations, 2) navigating, by attempting to grab a nearby hold, which may be used both to confirm the true reachability, as well as to traverse the environment, and 3) simulated trajectories, modeled by particle filtering (Sequential Monte Carlo rollouts) over the present surface. The first two tools are specified by the task, and the third is the central mechanism of ADP.

\paragraph{Visual Search}

As the fovea is moved across the map, precise visual information about the location (or absence of) holds is integrated into the energy surface. Specifically, regions absent of any hold are set to high energy values (indicating a vanishing probability the final path will land in this area), while the energy around regions where holds are discovered is reduced. However, the locations of holds themselves aren't sufficient to infer path passability, since one hold must be reachable from the other to allow traversal. 

Dynamical prospection supports this function.

\paragraph{Dynamical Prospection}

To support the learning of an energy surface suitable for navigation, we model prospection as parallel stochastic simulations of possible trajectories from the agent's present location. Particle dynamics sample a successor location from the present energy landscape, making hops to lower energy regions (where the agent already believes its path is likely to fall) more likely. 

Rollouts run in a latent world model, allowing simulation of paths that are not currently in the field of view. However, uncertainty about the locations and accessibility of holds in the model results in sampling variance and constraints to trajectory distance.

Central to ADP is the idea that simulated particle trajectories are most useful to an agent when governed by dynamics reflecting the agent-environment system's characteristics of interaction---both transition dynamics, and goal-seeking preferences. Specifically, we model particles with constraints imposed by a simple intuitive physics: momentum and a distance-aware sample filter. Particle momentum reduces most quickly when moving up an energy gradient, and more slowly when descending, resulting in longer rollouts during descent. The sample filter limits consideration for a particle's successor location to an approximate reach-length radius, thus ensuring particle dynamics parallel the agent's own ability to traverse the landscape. 

The sampled trajectories are sufficient to determine three effects which control the agent's internal representation and behavior:

\begin{enumerate}
    \itemsep -0.3em
    \item Particles update the underlying energy at each sampled location, as a function of their terminal energy.
    \item Because the agent expects to move through a low energy path, trajectories passing through high energy areas produce prediction errors. The location of these errors are used to attract visual attention (moves of the fovea), which generate observations to resolve this ambiguity.
    \item The direction of the first step of each rollout determines confidence in the next (navigational) move. When directional variance drops below a threshold (parameterizing greediness), the agent attempts to reach in the consensus direction.
\end{enumerate}

We hypothesized that, together, our agent model (ADP-Agent) would exhibit behaviors useful to solving the pathfinding task such as: greedy exploration of direct paths to goal, focusing visual search on optimistic but still ambiguous candidate path locations, and dynamic planning demonstrated by iterative use of visual search and hold traversal. Altogether, we expected the proposed model to be capable of solving maps with similar difficulty to those solvable by humans. A detailed description of the model implementation follows.

\subsection{Model Details}
\label{sec:model_details}

\paragraph{Agent Task Paradigm}

The agent task paradigm was modeled to maximize consistency with the problem posed to human participants, while abstracting away low-level motor dynamics like controlling a cursor during click and drag. 

Agent state is a tuple $ ( X_{agent}, X_{fovea}, E ) $, where $ X_{agent} \in \mathbb{R}^2 $ is the agent's location in the map, $ X_{fovea} \in \mathbb{R}^2 $ is the agent's fovea location (which determines the position of the spotlight), and $E$ is the internal model of the map as an energy landscape.

On each time step, the agent receives an observation from the local area around its fovea, which includes the positions of all holds in the map within a fixed foveal radius. The agent then chooses an action, composed of the next position for both the agent and the fovea: $A_t = ( X_{agent}(t+1), X_{fovea}(t+1) ) $. The agent need not move itself nor its fovea on every time step. The environment updates in response to the chosen action by 1) moving the agent location to $X_{agent}(t+1)$ if this location is reachable (distance within reach radius), and 2) moving the fovea to $X_{fovea}(t+1)$, taking multiple steps if fovea distance is greater than the maximum fovea velocity. If the agent's new position lies within a goal, the trial is completed successfully.


\paragraph{ADP-Agent}
\label{sec:agent_internal_representation}

ADP-Agent is instantiated with an energy landscape represented as a 2D matrix or raster $E^{W \times H}$ where each $e_{xy} \in [0, 1]$ represents the energy at that point in the landscape. $W$ and $H$ are parameters specifying the resolution of the agent's energy landscape. We separately define a distance-based energy floor, $E_{floor}$, calculated as the Euclidean distance to the closest goal location. $E$ is initialized to $E(t_0) = E_{floor} + C$ where $C$ is a constant.

We define the following additional parameters influencing various aspects of agent behavior:

\begin{itemize}
    \itemsep -0.3em 
    \item $k$: Number of particles to emit per step
    \item $\tau$: Softmax temperature for particle location sampling
    \item Particle mass $m$: Inverse of rate at which we reduce particle momentum during rollout
    \item $\alpha$: Learning rate for energy updates
    \item Move consensus threshold $\eta$: Percentage of first particle steps landing on the same hold required to attempt a move
    \item $d$: Energy decay rate (towards initial initial energy $E(t_0)$)
\end{itemize}

On each time step, ADP-Agent performs $k$ particle rollouts, instantiating each at the agent's present location $X_{agent}$\footnote{Alternative strategies were also explored, including sampling an origin from the landscape, or alternating between the fovea location and the agent. The agent-origination strategy was most robust in our simulations.}. Rollouts are computed by considering only locations in the energy landscape within an approximate reach radius from the location of the particle (for convenience, we implement this as a circular binary mask centered at position $X$, with radius $r$: $\verb|Mask(X, r)|$), resulting in a candidate subset of the landscape $E_{c}$. Particle dynamics then follow a softmax, such that the next location is sampled as:

\begin{eqnarray}
X_{p,j+1} \sim p(X_{p,j+1}|X_{p, j}, E) = \frac{exp(-E_{c}/\tau)}{\sum_{i} exp(-E_{c,i}/\tau) }
\end{eqnarray}

Particles lose momentum as a function of the change in energy of the landscape: $p_{j,mnt} \longleftarrow p_{j-1,mnt} - \frac{E[X_{p,j}] - E[X_{p,j-1}]}{m} $, and dampened by the particle mass parameter, $m$. The rollout continues until the particle's momentum falls to 0 or below.

\paragraph{Learning}

After each particle $i$ terminates, the energy landscape is updated underneath each step of its trajectory $\pi_i = \{ X_{i0}, X_{i1}, ..., X_{in} \}$ by an approximate momentum-discounted learning rule based on the difference between the energy at each step $E[X_{ij}]$, and that at the terminal location of the rollout $E[X_{in}]$. 

\begin{eqnarray}
E \longleftarrow E + \alpha p_{ij,mnt} (E[X_n] - E[\verb|Mask|(X_{ij})])
\end{eqnarray}

This learning update serves to push the landscape energy towards the terminal energy as illustrated in Figure \ref{fig:1d_energy}.

Following all rollouts and updates, the landscape is multiplicatively decayed (by rate $d$) towards its initial conditions, and clipped to the interval $[E_{floor}, 1]$ after each step: 

\begin{eqnarray}
E \longleftarrow clip(E + d (E(t_0) - E), E_{floor}, 1)
\end{eqnarray}



To choose a new location for the fovea, an error map $\Psi$ is computed by summing the energy under every particle trajectory step. In this way, areas of high surprise (particle trajectories moving through high energy regions) can be efficiently calculated as a direct result of the rollout computation.


\begin{eqnarray}
X_{fovea}(t+1) = \argmax_{X} \sum \Psi[\verb|Mask(X)|]
\end{eqnarray}

The second component of action, $X_{agent}(t+1)$, is determined based on the uncertainty (entropy) of first step directions over all trajectories. We define the set of first step directions (for a particle batch) as $\Theta = \{ \theta_i \}_{i=1}^k $ where $\theta_i = arctan(\frac{p^i_{y,1} - p^i_{y,0}}{p^i_{x,1}-p^i_{x,0}}) $. We then calculate the variance, and if $Var(\Theta) < \eta $, the agent attempts to reach the hold upon which the plurality of its first steps ($p_{x,1}, p_{y,1}$) fall. 

Videos of sample agent runs can be found at the media page linked in the `Data Structure' section above, and Python code is available as a public repository.

\section{Results}

\subsection{Online (Human) Experiment Results}

\subsubsection{Prospection via Attention Distance}

\begin{figure}
\begin{center}
\includegraphics[width=\linewidth]{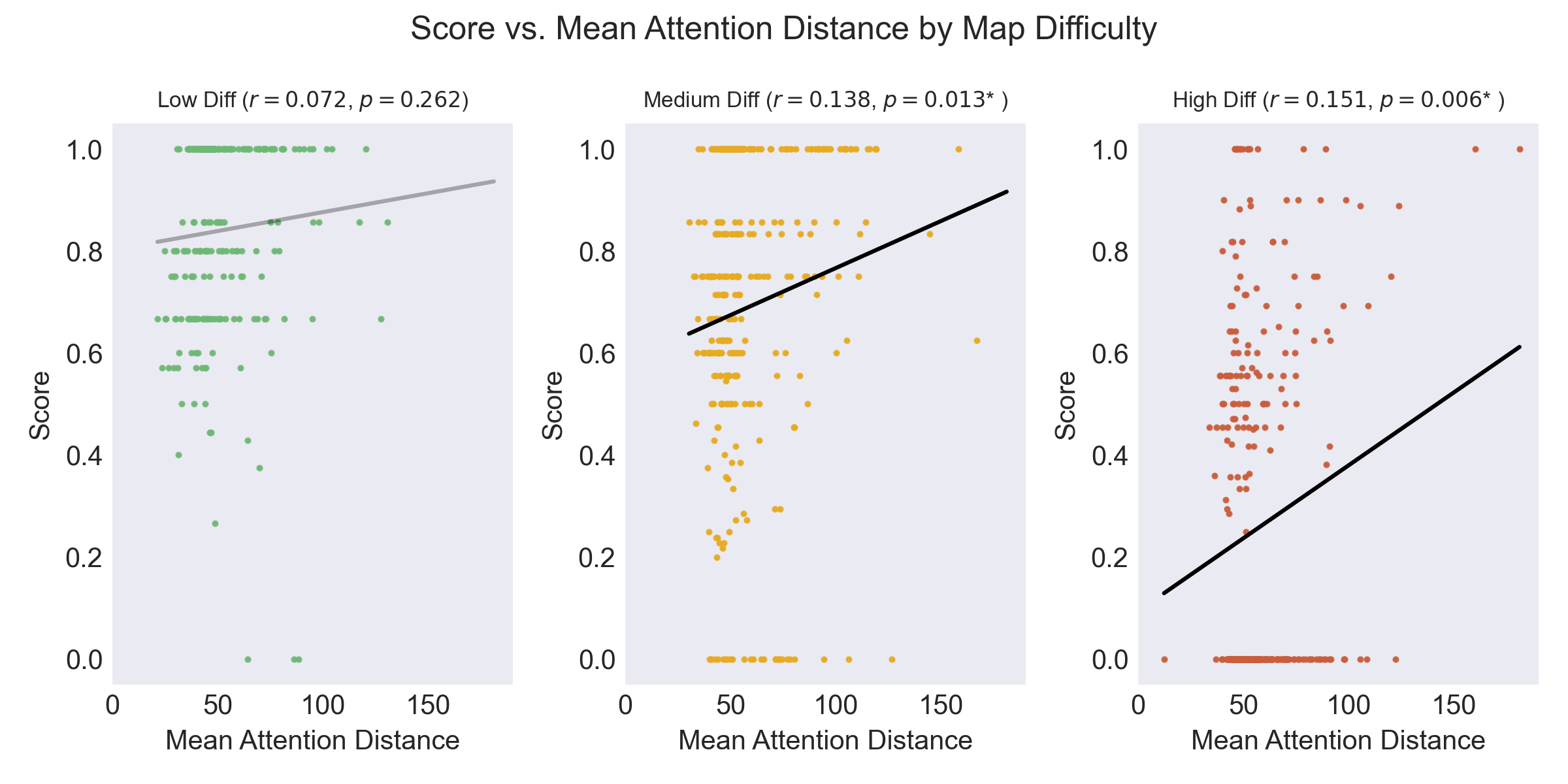}
\caption{Regressions of trial-wise score versus mean attention distance. An increasingly strong positive correlation is seen as map difficulty increases.} \label{fig:att_dist_regressions}
\end{center}
\end{figure}

We investigated both distributional and temporal characteristics of attention distance. At the trial level, we found a positive correlation between the mean of attention distance and score across all three difficulty levels. This relationship is statistically significant for medium and high-difficulty maps, but not for low-difficulty maps (see Figure \ref{fig:att_dist_regressions} for details).

\begin{figure}[t]
\begin{center}
\includegraphics[width=\linewidth]{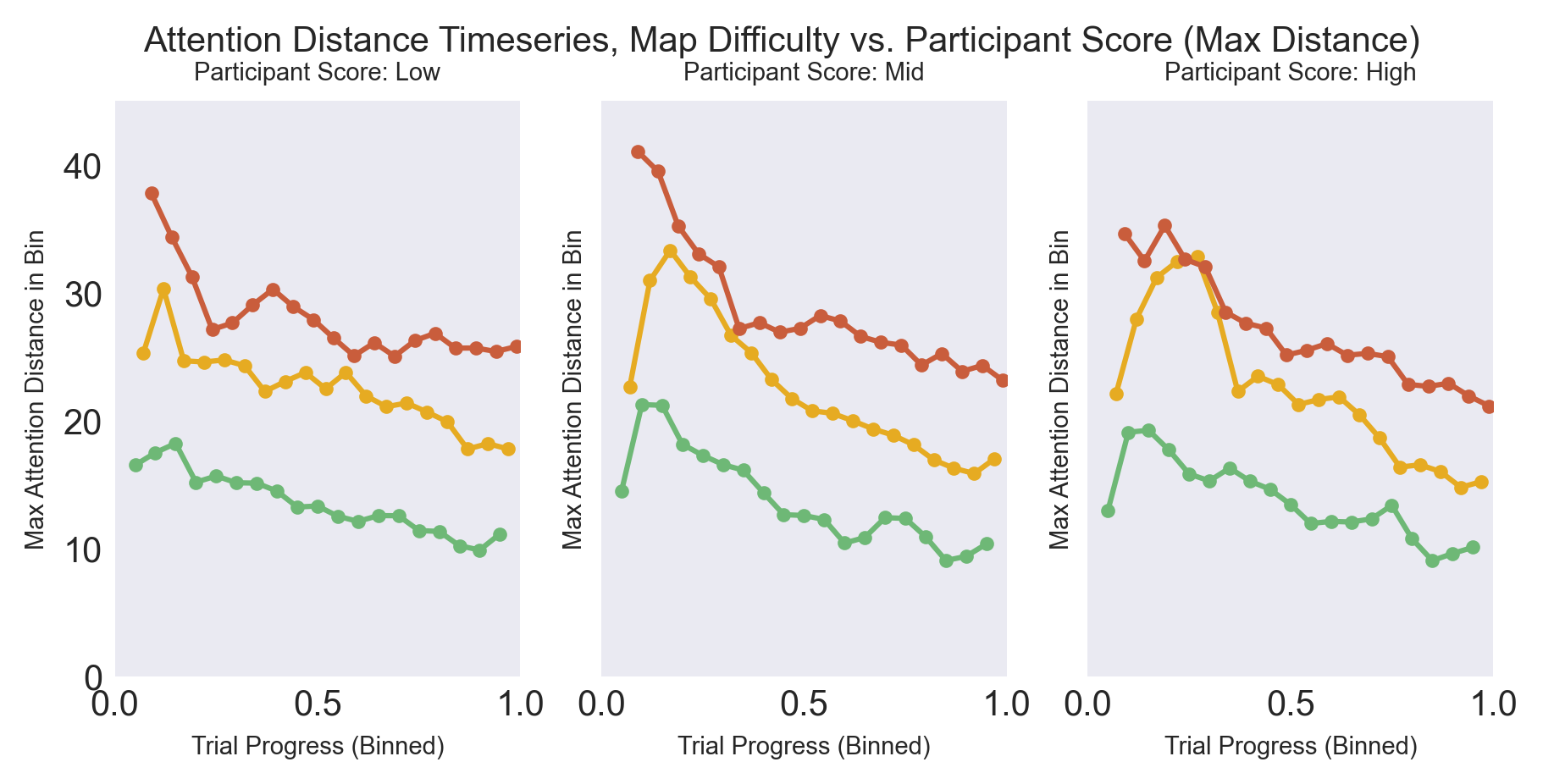}
\caption{Binned longitudinal max attention distance across map difficulty (green, orange, and red line series indicating low, medium and high difficulty), and participant performance groups (left, middle, and right charts). We observe consistent declining trends across each trial duration as exploration reduces during navigation (exploitation).} \label{fig:attention_timeseries_3}
\end{center}
\end{figure}

To identify patterns in temporal attention data, we computed maximum attention distance binned based on progress through trial, which allowed us to standardize longitudinal data across trials of varying duration. As shown in Figure \ref{fig:attention_timeseries_3}, we found a general trend of reducing distance as trials progress, as well as a positive relationship between distance and map difficulty. The downward trend was shallowest for the lowest-performing participant segment. 

\begin{figure}
\begin{center}
\includegraphics[width=\linewidth]{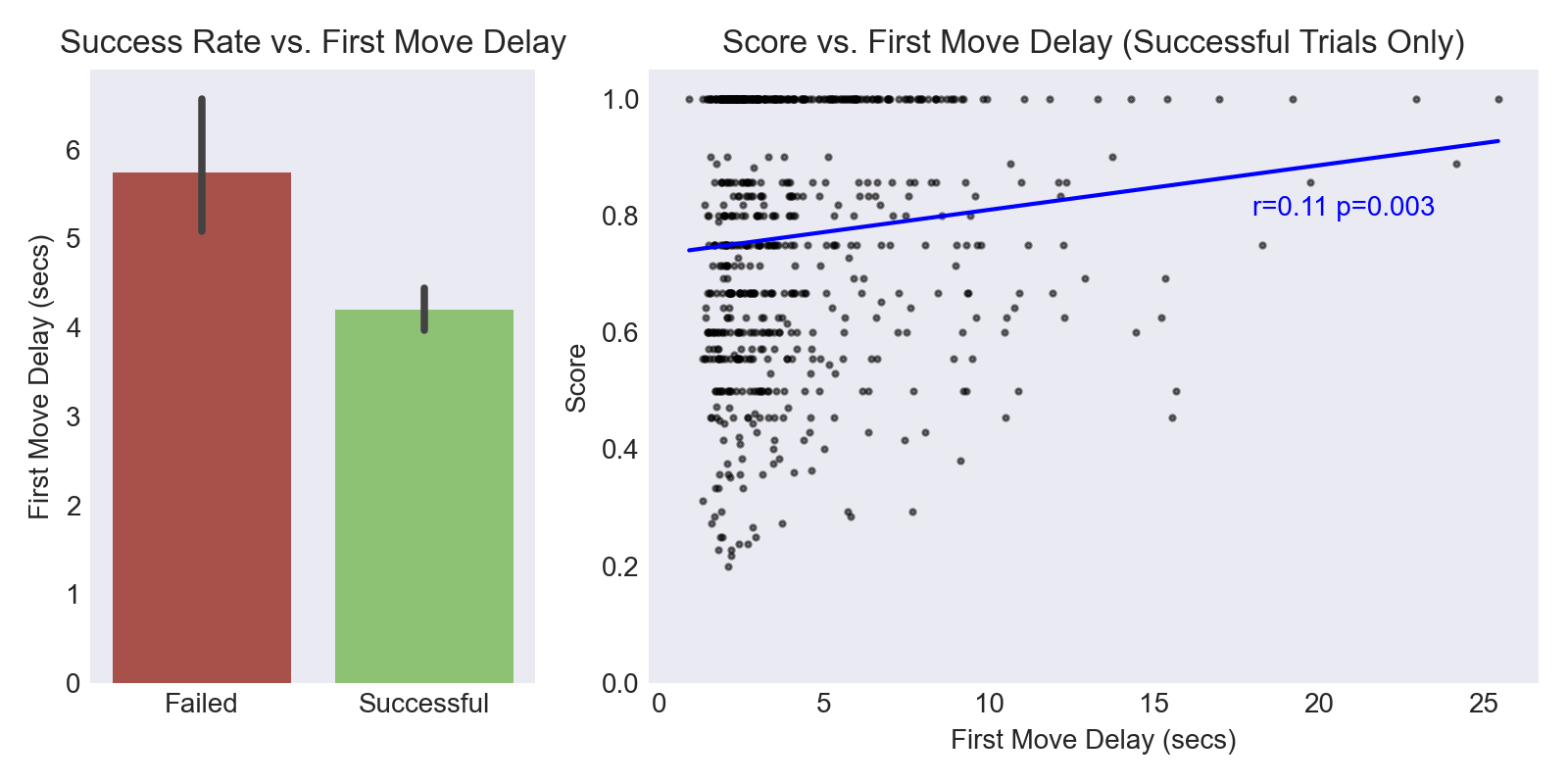}
\caption{Left: Trial-wise success versus delay before first move (in seconds). Right: Regression of trial score versus delay, among successful trials.} \label{fig:success_vs_delay}
\end{center}
\end{figure}

As another perspective on prospective and exploratory behavior, we analyzed the delay prior to first move. We found that on trials with longer delays (wherein participants explored the landscape for longer prior to navigating to their first hold) success rate overall was lower (see Figure \ref{fig:success_vs_delay}). However, when looking only at successful trials, we found a statistically significant association between delay and trial score. 





\subsection{Simulation Results}

Simulations were run using the same maps and task constraints as those used in the online experiment. 81 simulations were run on each map, with identically instantiated agents. We compared four primary outputs of simulation runs with the results from our online experiment: success rate, distribution of goal reached (for maps with multiple reachable goals), duration distribution, and spatial attention distribution. 

\begin{figure}
\begin{center}
\includegraphics[width=\linewidth]{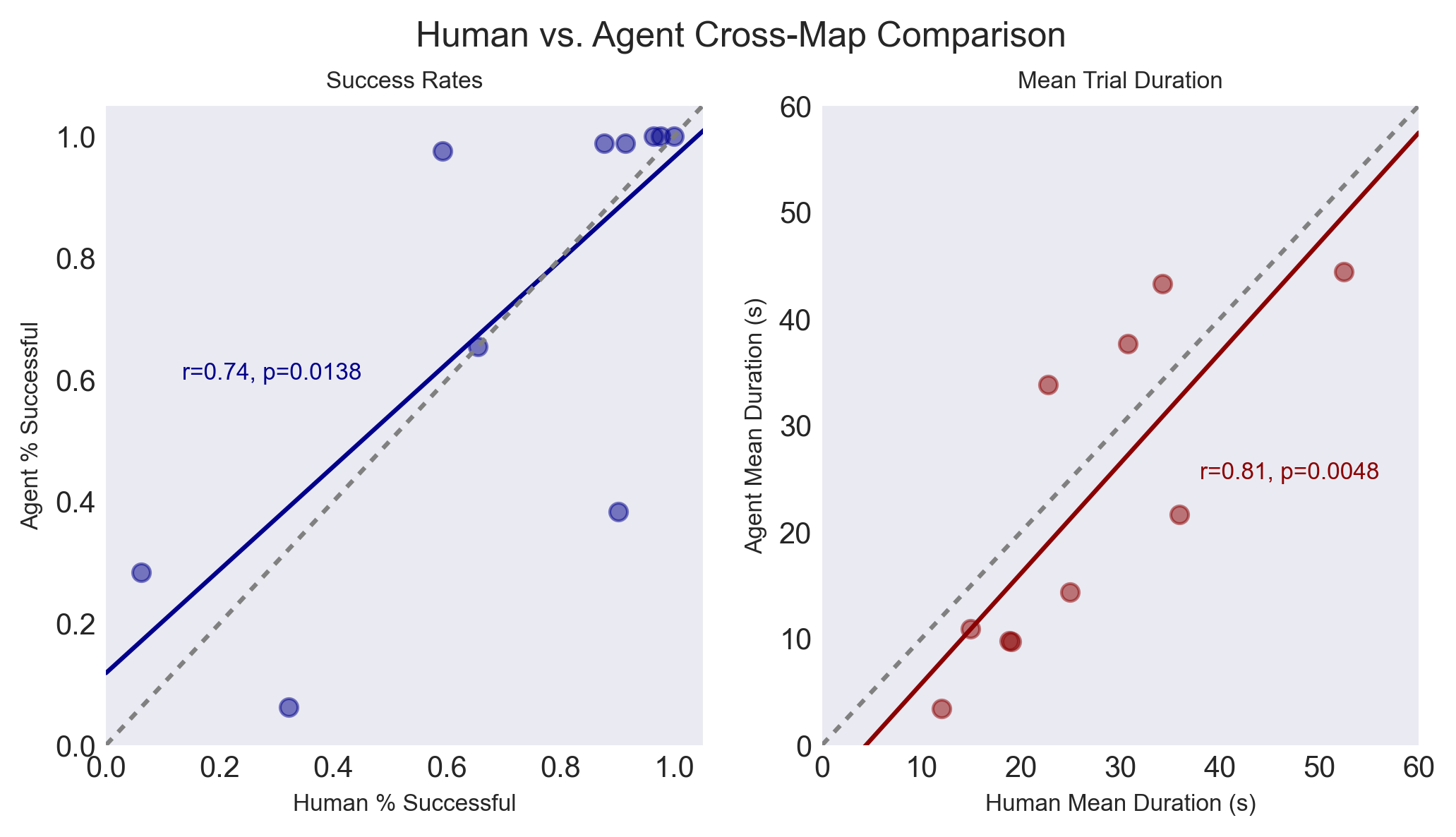}
\caption{Comparison of human and agent success rates (left) and trial durations (right) across all maps.} \label{fig:hum_agent_scatter_summary}
\end{center}
\end{figure}

Our results show that all maps can be successfully solved by ADP-Agent, and that success rates are well correlated with that of human participants (see Figure \ref{fig:hum_agent_scatter_summary}). While some maps showed similar goal distributions, indicating related goal choice dynamics, a minority showed inversed preferences (e.g. maps 2 \& 11). Trial duration was also highly correlated (Pearson-$r=0.81$, $p<0.005$). 

\begin{figure}[t!]
\includegraphics[width=\linewidth]{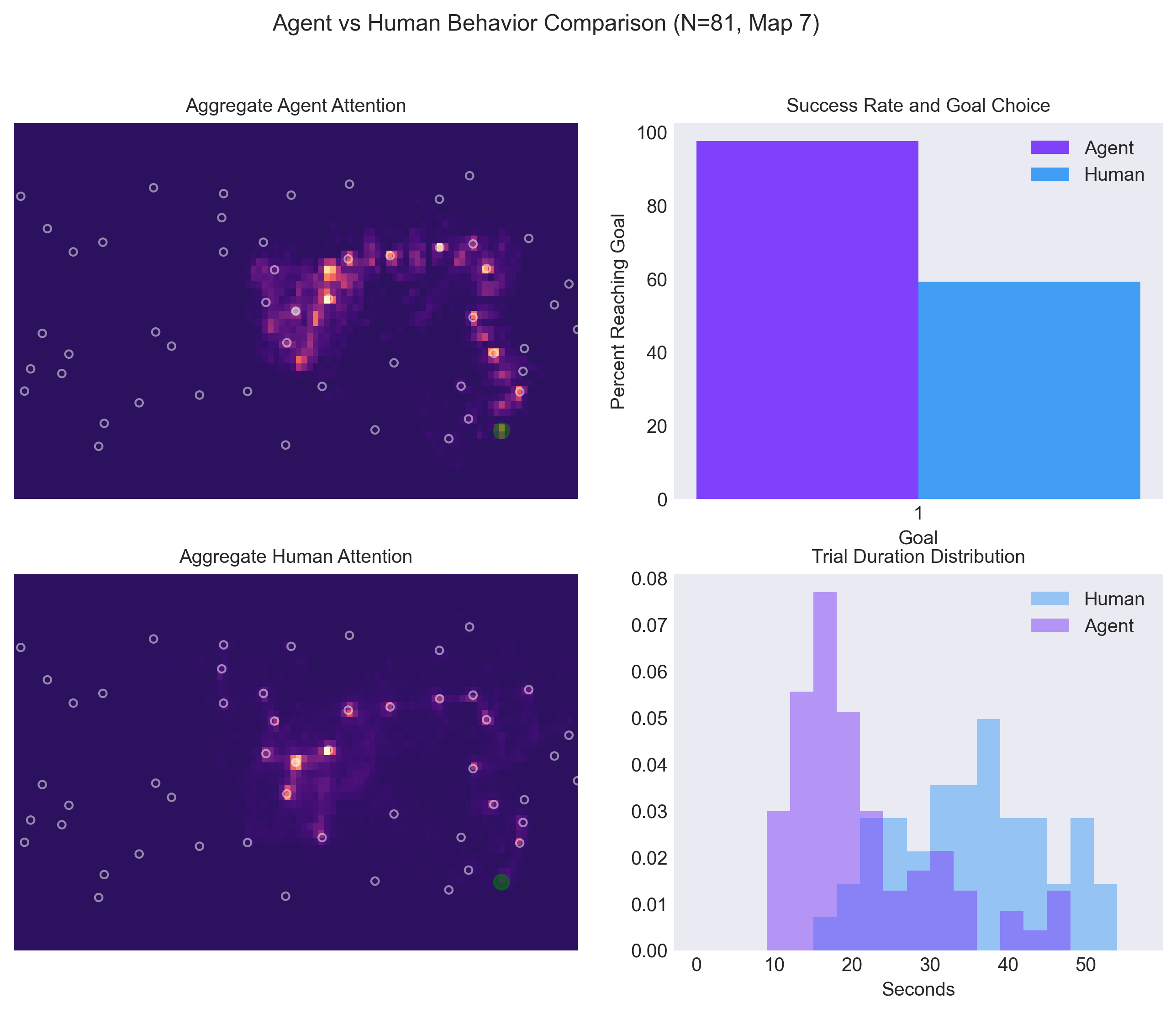}
\caption{Human vs. agent comparison (map 7). Left: aggregate attention outputs from all runs. Top right: goal reached success rate. Bottom right: run duration distribution. Agent simulations solved map 7 more quickly (in simulated time), and more consistently than humans. } \label{fig:sim_agent_comp_sample}
\end{figure}

\section{Discussion}
\label{sec:discussion}

Our behavioral results showing an increasingly significant correlation between mean attention distance and trial-wise score can be interpreted as the value of exploratory distal attention. Though weaker for low difficulty maps, which we might expect since a greedy no-look-ahead policy was still effective in these cases, participants (and agents) could easily get stuck in dead ends if they didn't confirm connectivity prior to movement down a path.

The temporal trend seen in attention distance, as well as the relationship between first move delay and score, suggest that participants had to balance visual exploration with navigation in order to succeed at our task. With failed trials filtered out (as was done for the delay versus score regression in Figure \ref{fig:success_vs_delay}), score can be seen as a proxy for path efficiency. While participants spending too long exploring prior to navigation were less likely to succeed, for those who were successful, exploratory behavior prior to committing to a spatial direction was predictive of path efficiency. 

In the following sections we discuss simulation results and their comparison with human behavioral data. While quantitative comparisons of path choice, trial duration, and success rate offer some validation that simulated agents generate attributes that are, in aggregate, consistent with human planners, we can also derive insights from qualitative analysis of behaviors seen in single simulation runs. 

\paragraph{Epistemic Value via Prospection}

As particles move through previously observed terrain (across high contrast or ``well-worn'' trajectory segments in the landscape), they follow predictable paths. However, when moving into unexplored terrain, the sampling dynamics generate splits and radiating branches guided only loosely by the underlying distance-based floor. Trajectories venturing into these higher energy regions produce large areas of prediction error, which suggest epistemic richness given a combination of high expected surprisal, and high path salience. ADP-Agent's foveal policy, which moves attention to the area containing maximum prediction error on the prior time step, therefore serves to expand the peripheries of the known landscape where it is most likely to yield paths to a goal. 

We also find dynamics in which particles ``jump off'' a path of observed holds influenced by an energy well from a nearby goal---even when these jumps take particles into unobserved regions. These trajectories might be thought of as optimistic shortcuts, and the high prediction errors they produce attracts visual attention to confirm or deny the hypothesis of path connectivity.

\paragraph{Attention \& Surprise}

A feature shared by human and agent attention is a focus on holds that are close to the reach limit, but not in fact reachable. Though distal scans of these connections may be assessed as passable (by humans, as well as by optimistic particle trajectories), upon arriving at the hold, a failed reach attempt prompts subsequent attempts, or consideration of alternative nearby paths. 

Other map attributes that are seen to attract attention across both simulations and behavioral data are symmetrical forks (in which two holds appear to lie on similarly direct paths to goal), and other regions of uncertainty caused by competing candidate trajectories. ADP-Agent fixates on these regions during increasingly long rollouts until a confidence threshold is reached\footnote{This is a dynamic consistent with theories of evidence accumulation, e.g. \cite{lee2004evidence}.}. In general, our model appears to leverage the parallelized nature of prospective simulations, with serially executed attentional movements supporting uncertainty reduction at the areas of highest error.



\paragraph{Search Depth \& Backtracking}

A common challenge in complex planning problems is the optimization of search depth, to avoid actions leading to dead ends. Backtracking was common in both human and agent simulations, especially in high difficulty maps including direct but ultimately disconnected paths. Forward search depth is modulated by ADP-Agent's particle mass and move consensus threshold parameters, which affect the length of trajectories, and navigational greediness, respectively. Empirical optimization of these parameters to a specific map (via grid search) was usually sufficient to achieve 100\% success rate on even the most challenging problems.

\subsection{Limitations \& Future Work}

The model presented here lacks several features inherent to human pathfinding that may limit its ability to predict and explain behavior. First, ADP-Agent is unable to generalize or treat clusters of holds or path segments as more abstract units. For example, while human participants likely perceive a sequence of closely positioned holds as a single passable route affording traversal from start to end, the landscape in our model independently represents an energy well around each hold. Secondly, some human attentional data appeared consistent with bi-directional planning (a well-known dimensionality reduction strategy long studied in psychology and artificial intelligence, e.g. \cite{pohl1971bi}), especially when confronted with challenging problems. In contrast, ADP-Agent's attention was seen to progress roughly monotonically towards goal locations driven by errors on the periphery of the observed landscape. Experimenting with particle emissions strategies that support inverse rollouts from goal locations may begin to address this limitation. 

\section{Conclusion}

In this work, we propose a computational model of visual exploration and navigation during pathfinding in a partially observable and uncertain environment. Results from simulations show that agents can successfully solve the task by minimizing prediction error generated by stochastic particle rollouts across a learned energy landscape. Behavioral data from our online experiment provides further insight into the range of strategies employed, and dynamics of prospective visual search during pathfinding. 

\section{Acknowledgements}

This material is based upon work supported by the National Science Foundation Graduate Research Fellowship Program under Grant No. 2019236659.

\footnotesize
\bibliographystyle{apalike}
\bibliography{bib} 

\begin{thebibliography}{}

\bibitem[Arnold et~al., 2016]{arnold2016mental}
Arnold, A.~E., Iaria, G., and Ekstrom, A.~D. (2016).
\newblock Mental simulation of routes during navigation involves adaptive
  temporal compression.
\newblock {\em Cognition}, 157:14--23.

\bibitem[Chersi et~al., 2013]{chersi2013mental}
Chersi, F., Donnarumma, F., and Pezzulo, G. (2013).
\newblock Mental imagery in the navigation domain: a computational model of
  sensory-motor simulation mechanisms.
\newblock {\em Adaptive Behavior}, 21(4):251--262.

\bibitem[Colorni et~al., 1991]{colorni1991distributed}
Colorni, A., Dorigo, M., Maniezzo, V., et~al. (1991).
\newblock Distributed optimization by ant colonies.
\newblock In {\em Proceedings of the first European conference on artificial
  life}, volume 142, pages 134--142. Paris, France.

\bibitem[Cos et~al., 2021]{cos2021changes}
Cos, I., Pezzulo, G., and Cisek, P. (2021).
\newblock Changes of mind after movement onset: a motor-state dependent
  decision-making process.
\newblock {\em bioRxiv}.

\bibitem[Friston et~al., 2017]{friston2017active}
Friston, K., FitzGerald, T., Rigoli, F., Schwartenbeck, P., and Pezzulo, G.
  (2017).
\newblock Active inference: a process theory.
\newblock {\em Neural computation}, 29(1):1--49.

\bibitem[Kaplan and Friston, 2018]{kaplan2018planning}
Kaplan, R. and Friston, K.~J. (2018).
\newblock Planning and navigation as active inference.
\newblock {\em Biological cybernetics}, 112(4):323--343.

\bibitem[Lakshminarasimhan et~al., 2020]{lakshminarasimhan2020tracking}
Lakshminarasimhan, K.~J., Avila, E., Neyhart, E., DeAngelis, G.~C., Pitkow, X.,
  and Angelaki, D.~E. (2020).
\newblock Tracking the mind’s eye: Primate gaze behavior during virtual
  visuomotor navigation reflects belief dynamics.
\newblock {\em Neuron}, 106(4):662--674.

\bibitem[Lee and Cummins, 2004]{lee2004evidence}
Lee, M.~D. and Cummins, T.~D. (2004).
\newblock Evidence accumulation in decision making: Unifying the “take the
  best” and the “rational” models.
\newblock {\em Psychonomic bulletin \& review}, 11(2):343--352.

\bibitem[Montello, 2005]{montello2005navigation}
Montello, D.~R. (2005).
\newblock {\em Navigation.}
\newblock Cambridge University Press.

\bibitem[Newell et~al., 1972]{newell1972human}
Newell, A., Simon, H.~A., et~al. (1972).
\newblock {\em Human problem solving}, volume 104.
\newblock Prentice-hall Englewood Cliffs, NJ.

\bibitem[Pezzulo et~al., 2019]{pezzulo2019planning}
Pezzulo, G., Donnarumma, F., Maisto, D., and Stoianov, I. (2019).
\newblock Planning at decision time and in the background during spatial
  navigation.
\newblock {\em Current opinion in behavioral sciences}, 29:69--76.

\bibitem[Pohl, 1971]{pohl1971bi}
Pohl, I. (1971).
\newblock Bi-directional search.
\newblock {\em Machine intelligence}, 6:127--140.

\bibitem[Silver and Veness, 2010]{silver2010monte}
Silver, D. and Veness, J. (2010).
\newblock Monte-carlo planning in large pomdps.
\newblock Neural Information Processing Systems.

\bibitem[Trianni and Tuci, 2009]{trianni2009swarm}
Trianni, V. and Tuci, E. (2009).
\newblock Swarm cognition and artificial life.
\newblock In {\em European Conference on Artificial Life}, pages 270--277.
  Springer.

\bibitem[Tschantz et~al., 2020]{tschantz2020scaling}
Tschantz, A., Baltieri, M., Seth, A.~K., and Buckley, C.~L. (2020).
\newblock Scaling active inference.
\newblock In {\em 2020 International Joint Conference on Neural Networks
  (IJCNN)}, pages 1--8. IEEE.

\bibitem[Yoon et~al., 2018]{yoon2018control}
Yoon, T., Geary, R.~B., Ahmed, A.~A., and Shadmehr, R. (2018).
\newblock Control of movement vigor and decision making during foraging.
\newblock {\em Proceedings of the National Academy of Sciences},
  115(44):E10476--E10485.

\end{thebibliography}

\end{document}